\documentclass[letterpaper, 10 pt, conference]{ieeeconf}                                                          
\IEEEoverridecommandlockouts                              
\usepackage{graphics} 
\usepackage{epsfig} 

\usepackage{times} 
\usepackage{gensymb}
\usepackage{cite}

\usepackage{amsmath} 
\usepackage{amssymb}  

\usepackage{amsthm}

\graphicspath{ {images/} }

\usepackage{algorithm}
\usepackage{algpseudocode}

\theoremstyle{remark}

\makeatletter
\newcommand{\removelatexerror}{\let\@latex@error\@gobble}
\makeatother

\title{\LARGE \bf
Below Horizon Aircraft Detection Using Deep Learning for Vision-Based Sense and Avoid
}

\author{Jasmin James,  Jason J. Ford and Timothy L. Molloy
\thanks{The authors are with the School of Electrical Engineering and Computer Science, Queensland University of Technology, 2 George St, Brisbane QLD, 4000 Australia. {\tt\small jasmin.james@connect.qut.edu.au, j2.ford@qut.edu.au, t.molloy@qut.edu.au}. This work was supported by funding from the Australian Research Council Centre of Excellence CE140100016 in Robotic Vision.}%
}

\begin{document}

\maketitle
\thispagestyle{empty}
\pagestyle{empty}

\begin{abstract}
Commercial operation of unmanned aerial vehicles (UAVs) would benefit from an onboard ability to  sense and avoid (SAA) potential mid-air collision threats. 
In this paper we present a new approach for detection of aircraft below the horizon. We  address some of the  challenges faced by existing vision-based SAA methods such as  detecting stationary aircraft  (that have no relative motion to the background), rejecting moving ground vehicles, and simultaneous detection of  multiple aircraft. 
We propose a multi-stage, vision-based aircraft detection system which utilises deep learning to produce candidate aircraft that we track over time.
We evaluate the performance of our proposed system on real flight data where we demonstrate detection ranges comparable to the state of the art with the additional capability of  detecting stationary aircraft, rejecting moving ground vehicles, and tracking multiple aircraft.
\end{abstract}

\section{Introduction}
The unmanned aerial vehicles (UAV) market is expected to grow to US \$51.85 billion by 2025 from US \$11.45 billion in 2016, with predicted growth in many important applications including: military and defense, retail, media, agriculture, industrial, law enforcement, construction, mining, oil and gas, telecommunications and many others \cite{Researchandmarkets}. 
Safe operation of an aircraft in the national airspace has historically assumed  human pilots' ability to visually see and avoid potential mid-air collision threats. The development of systems  capable of matching (and exceeding) the performance of human pilots is one of the key technical challenges hindering more routine, standard and flexible operation of UAVs in the national airspace \cite{Clothier2015}. 

Sense and avoid (SAA) refers to the implied regulatory requirement that UAVs be capable of sensing and avoiding potential collision threats with performance equivalent to that expected from a  human pilot.  Machine vision has been identified as a promising technology for the ``sense" aspect in small to medium sized UAVs, as vision sensors have size, weight, power and cost advantages over other sensing approaches such as radar \cite{Mcfadyen2016}. 

\begin{figure}[]
\begin{center}
\includegraphics[scale=0.5]{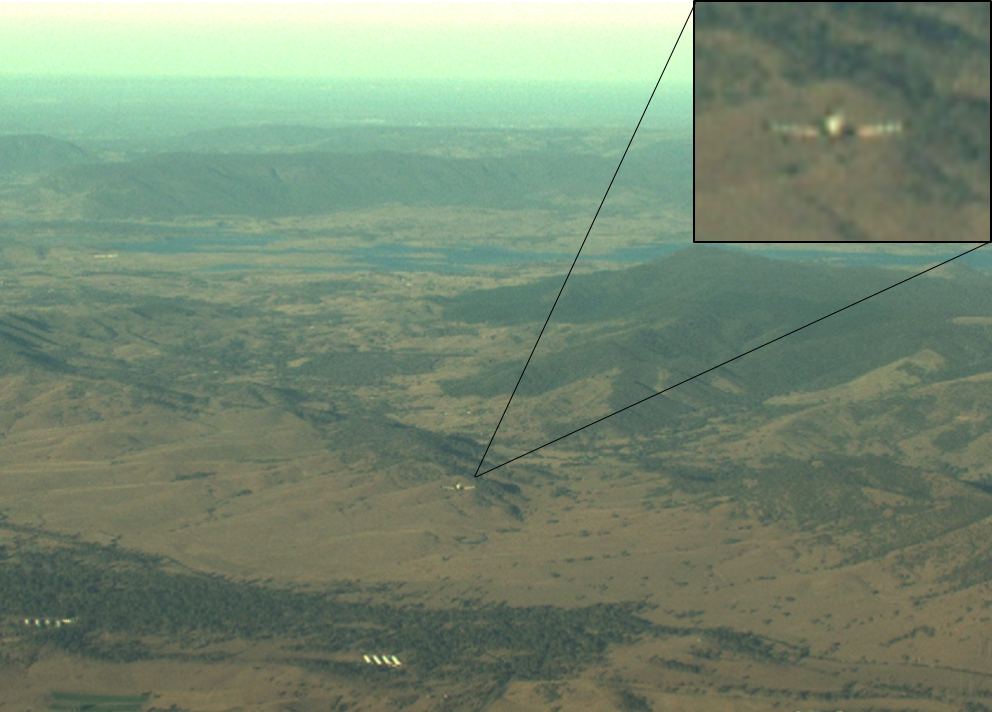}
  \caption{An example of an aircraft below the horizon amid ground clutter. This image  was captured in the Watts Bridge area of Queensland, Australia.  }
\label{fig:stationaryExample}
\end{center}
\end{figure}

Numerous approaches for  long range, fixed-wing,  vision-based aircraft detection have been presented in the literature. Progress in detecting aircraft above the horizon has been steady with advances in reliable systems that can detect at ranges approaching those of human pilots \cite{Nussberger2014,Lai2013,zarandy2015novel,JamesRAL}.
Detection of aircraft below the horizon presents new challenges as the environments are complex and cluttered as seen in Figure \ref{fig:stationaryExample}.

\begin{figure}[]
\begin{center}
\includegraphics[scale=0.8]{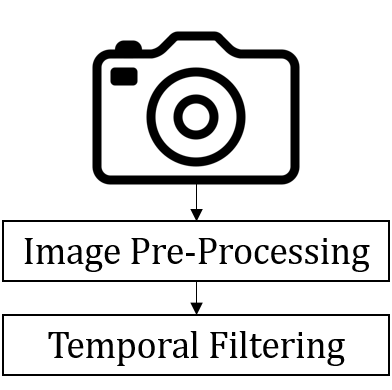}
  \caption{The key stages of the vision-based aircraft detection pipeline.}
\label{fig:stages}
\end{center}
\end{figure}

\begin{figure}[]
\begin{center}
\includegraphics[scale=0.5]{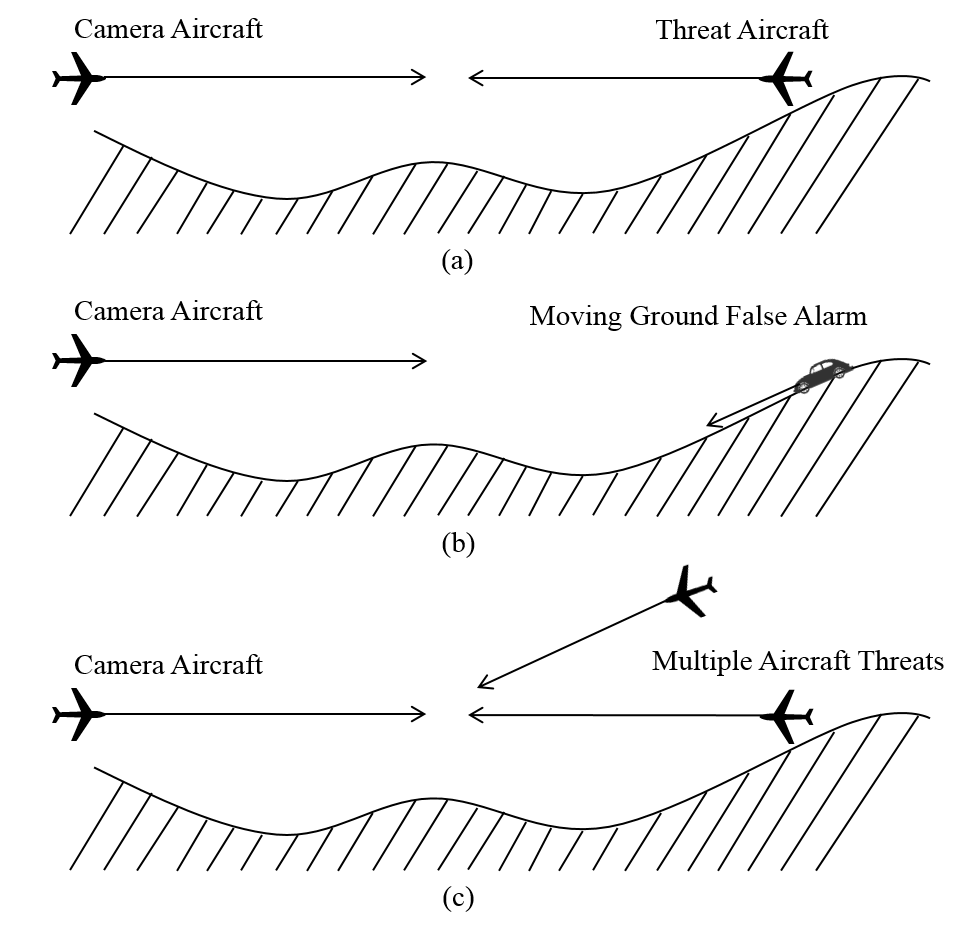}
  \caption{Some challenging scenarios: (a) detection of a true collision course encounter that may appear stationary with respect to ground clutter, (b) rejection of a moving ground vehicle false alarm and (c) detection of multiple-aircraft  threats (reproduced from \cite{Nussberger2014,molloy17}).}
\label{fig:challenge}
\end{center}
\end{figure}

 The most successful approaches for  below horizon vision-based aircraft detection exploit a multi-stage detection pipeline of image pre-processing and temporal filtering as seen in Figure \ref{fig:stages}  \cite{molloy17,Nussberger2014,zarandy2015novel}. 
The image pre-processing stage aims to extract potential aircraft in an image. The dominant approach is to use  frame differencing (or background subtraction) to extract potential aircraft that have apparent motion with respect to the background  \cite{Nussberger2014,molloy17, zarandy2015novel}.  
The temporal filtering stage aims to track these potential aircraft in an image sequence. In \cite{Nussberger2014} an extended Kalman filter is used and a ``valid track'' of the aircraft (where the aircraft is consistently detected) is declared at an average detection range of $1747$m with an average of $4$ false alarms over their tested image sequences.  In \cite{molloy17} a HMM filter is used for detection of aircraft below the horizon with an average detection range of $1890$m and no false alarms.
Whilst these multi-stage frame differencing approaches are considered to be the state of the art there are some key limitations that should be considered, see Figure \ref{fig:challenge}. A fundamental problem with frame differencing approaches is that threat aircraft on a true collision course may have small or no apparent motion with respect to the background \cite{zarandy2015novel,molloy17}. Moreover, the majority of these approaches have been adapted from detecting moving ground vehicles, and hence struggle to reject this type of false alarm \cite{molloy17}. Finally, the detection system of \cite{molloy17} is for a single aircraft and does not detect multiple aircraft threats. In this paper we propose a system to address these concerns  by exploiting  aircraft visual appearance rather than relying on apparent motion with respect to the background.

To exploit the visual appearance of aircraft, various machine learning  \cite{Dey2011} and deep learning \cite{Rozantsev2014,Hwang2018} approaches have  been investigated for vision-based sense and avoid. 
 In  \cite{Dey2011} a multi-stage detection  pipeline is proposed which used a support vector machine (SVM) to exploit aircraft visual features. An average  detection rate of $98\%$ of the tested images with aircraft present out to $8$km is achieved with a false alarm rate of 1 every 50 frames.   In \cite{Rozantsev2014} they propose an approach which uses spatio-temporal image cubes for classification. The authors report an average precision of $75\%$ on their UAV dataset and $79\%$ on their aircraft dataset.  In \cite{Hwang2018} the authors propose a deep CNN which is able to detect aircraft and achieved a $83\%$ detection rate on the tested images with aircraft present. Recently \cite{JamesRAL} used a deep CNN fused with morphological processing to detect aircraft above the horizon with a mean detection range of  $2527$m and no false alarms. Despite their potential for below horizon aircraft detection, prior machine and deep learning approaches have under achieved in detection ranges, false alarm rates or have solely been demonstrated for above horizon aircraft detection.
 
The key contribution of this paper is the proposal of a new system for below horizon vision-based aircraft detection that addresses some of the key challenges faced by the current state of the art  including:
\begin{itemize}
    \item Detection of stationary aircraft (that have no relative motion to the background);
    \item Rejection of moving ground vehicles; and,
    \item Simultaneous detection of multiple aircraft.
\end{itemize}
Our proposed system maintains the structure of Figure \ref{fig:stages} but in contrast to frame differencing exploits aircraft visual appearance for  image pre-processing and a simple temporal filtering stage capable of tracking multiple aircraft. 

The rest of this paper is structured as follows.  In Section \ref{sec:app} we describe our proposed approach for learning to detect aircraft below the horizon. In Section \ref{sec:testData} we describe our testing data.  In Section \ref{sec:results} we  experimentally investigate the performance of our proposed system. In Section \ref{sec:disc} we discuss our proposed system and its limitations. Finally, we provide some conclusions in Section \ref{sec:conc}.

\section{Proposed System}\label{sec:app}
In this section we describe our proposed  multi-stage system (see Figure \ref{fig:stages}) for below horizon vision-based aircraft detection. We first describe our  deep learning image pre-processing stage (including our network and training process). We then describe our temporal filtering stage. 

\subsection{Image Pre-Processing Stage}
We propose using deep learning to extract potential aircraft candidates in an image. 
Our proposed network is a variation of the SegNet architecture first presented in  \cite{Badrinarayanan} for pixel-wise segmentation and more recently used in \cite{JamesRAL} for detection of aircraft above the horizon. The output of this network is an image where each pixel is classified as either ``aircraft" or ``background" (the potential ``aircraft" pixels can then be tracked in the temporal filtering stage). We now briefly discuss our network structure and training process.

\begin{figure}[]
\begin{center}
\includegraphics[scale=0.4]{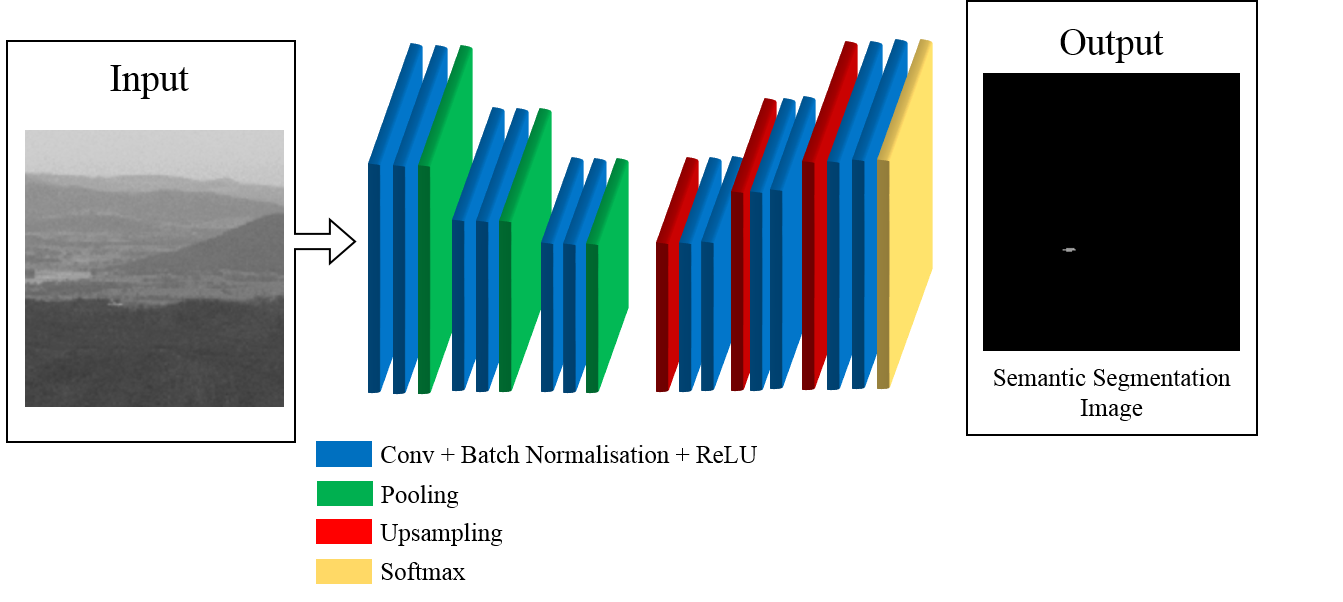}
  \caption{An illustration of proposed network architecture for below horizon aircraft detection.  The input to the system is a greyscale image, there are three encoders/decoder layers which are fed into a softmax layer for pixel-wise classification.}
\label{fig:network}
\end{center}
\end{figure}

\subsubsection{Network Architecture}
Our proposed architecture consists of an encoder network, a corresponding decoder network and a pixel-wise classification layer. This network architecture is fully convolutional (i.e., it has no fully connected layers). We modified the SegNet proposed in \cite{Badrinarayanan} to have $3$ encoder/decoder layers as seen in Figure \ref{fig:network}. We found $3$ encoder layers to be sufficient in the sense that we could still learn key visual aircraft features even with our limited training data (a more complex network would require more data to train). 
Each encoder layer in our network performs 64 $3\times3\times1$ convolutions with a stride of $[1 \ 1]$. The convolution outputs are  batch normalised and an element-wise rectified-linear non-linearity (ReLu) is applied. The corresponding decoder upsamples its input feature maps using the memorized max pooling indices, see \cite{Badrinarayanan} for more details. The output of the final decoder layer is fed to a softmax layer which classifies each pixel independently into either ``aircraft" or ``background".

\subsubsection{Training  and Labelling}\label{subsec:train}
To train our network we utilised the greyscale image sequences from \cite{molloy17}. We used cases $11-63$  (cases $1-10$ were reserved for testing), as well as $21$  additional cases that were excluded for various reasons (aircraft going in and out of field of view, moving ground vehicles, etc.).
These image sequences are head-on, near collision course encounters between two manned aircraft: a Cessna 172R camera aircraft; and a Socata Trinidad target aircraft (with a wingspan of $9.97$m). The aircraft data was captured with an uncalibrated Basler Scout machine vision camera with a Navitar NMV-12m-23 lens mounted on the wing strut of the camera aircraft. The camera was configured to capture $1280 \times 960$ pixel frames in 8-bit Bayer mode at approximately 15 Hz. The camera aircraft was flown at a desired altitude of between 823m and 914m, and the target aircraft operated at a vertical separation greater than 60m below the sensor aircraft.  See \cite{molloy17} for more details of the flight experiments.

\begin{figure}[t]
\begin{center}
\includegraphics[scale=0.75]{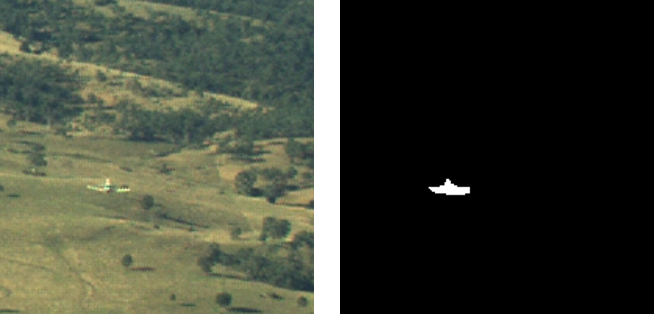}
  \caption{An example of a raw image (left) and its corresponding labelled image (right) where the white pixels are ``aircraft" and the black pixels are ``background". }
\label{fig:labComp}
\end{center}
\end{figure}

 From these image sequences we used the frames that featured distinguishable aircraft to create our training dataset. In total we used $17064$ images of target aircraft data.
Each training image was manually labelled and each pixel was classified as either aircraft or background. This was done in MATLAB for each frame by tracing around the aircraft using the function \textit{imfreehand}. To efficiently train our network we used a training image size of $200 \times 200$ and randomly cropped around the aircraft so that it was present in various locations in the images. An example of a labelled and cropped image is presented in Figure \ref{fig:labComp}.

To train our network, we weighted the loss function to balance the aircraft and background classes.
The  weights were initialised using the common `MSRA' weight initialization method \cite{He2015}. The optimisation algorithm used for training was stochastic gradient descent with momentum $0.9$, initial learn rate $0.001$, L2 regularisation $0.0005$ and max epochs $200$. Additionally, we used data augmentation to provide more training examples to the network, including random left/right reflection and random $X/Y$ translation of $\pm 10$ pixels. We implemented our network in MATLAB using the function ``segnetLayers''.   The implementation processed approximately 2 frames per second on a PC running Ubuntu $14.04$ with a $3.4$GHz Intel Core i7-6700 CPU and NVIDIA GeForce GTX 1070 GPU. We found that  with a suitably chosen threshold value of $0.999$, our network was successfully able to extract potential aircraft from the image sequences with a low number of false positives.


\subsection{Temporal Filtering}
We utilised a simple temporal filtering stage where we tracked potential aircraft over sequential frames.  Specifically, we checked  sequential frames for pixels classified as aircraft within a $10\times 10$ pixel region (from the centroid of the aircraft) and declared a detection when an aircraft was present for $W$ successive frames; we term this $W$ our window length.
Importantly, we highlight that because we are checking sequential images within a  region we did not require the images to be stabilised.


\section{Testing Data}\label{sec:testData}
In this section we describe the flight experiments that we used to evaluate the performance of our proposed system. 
\begin{table}[]
\centering
\renewcommand{\arraystretch}{1.3}
\caption{Summary of our testing data}
\label{tbl:dataDesc}
\begin{tabular}{|l|l|l|}
\hline
\normalsize Label           &  Collection Date (AEST)    & Pass Direction   \\ \hline
\normalsize T1 & April 8 2015  (13:41:30)   & SW\\
\normalsize T2 & April 8 2015  (13:51:23)    & SW\\
\normalsize T3 & April 8  2015  (13:56:13)  & NE\\
\normalsize T4 & April 8  2015  (14:00:38)   &  SW\\
\normalsize T5 & April 8 2015  (14:04:42)   & NE\\
\normalsize T6 & April 8  2015  (14:09:24)    & SW\\
\normalsize T7 & April 8  2015  (14:14:40)  & NE\\
\normalsize T8 & April 8 2015  (14:19:30)   &  SW\\
\normalsize T9 & April 8 2015  (14:23:54)  & NE\\
\normalsize T10 & April 8 2015  (14:28:15)   &  SW\\ 
\hline
\normalsize S1 & June 7 2016  (16:14:13)   & N/A\\
\normalsize S2 & June 7 2016  (13:22:53)    & N/A\\
\hline
\normalsize G1 & July 16 2015  (12:26:03)   & SW\\
\normalsize G2 & July 16 2015  (12:39:39)    & NE\\
\normalsize G3 & November 18 2015  (15:06:38)  & SE\\
\normalsize G4 & November 18 2015  (15:55:18)   &  NW\\
\hline
\normalsize M1 & November 19 2015  (11:40:50)   & NE\\
\hline
\end{tabular}
\end{table}

These experiments were conducted using the same aircraft and setup as described in \cite{molloy17} (and in  the  above  training and labelling section). 
We examined a variety of different image sequences to characterise our proposed system including:
\begin{enumerate}
    \item Head-on encounters;
    \item Stationary aircraft encounters;
    \item Moving ground vehicle encounters; and,
   \item  Multiple aircraft encounters.
\end{enumerate}
See Table \ref{tbl:dataDesc} for a summary of these flight experiments.  We now describe these image sequences in more detail.

\subsubsection{Head-on Encounters (T1-T10)}
We used cases $1-10$ of the head-on, near collision course encounters presented in \cite{molloy17}. 

\subsubsection{Stationary Aircraft Encounters (S1,S2)}
\begin{figure}[t]
\begin{center}
\includegraphics[scale=0.6]{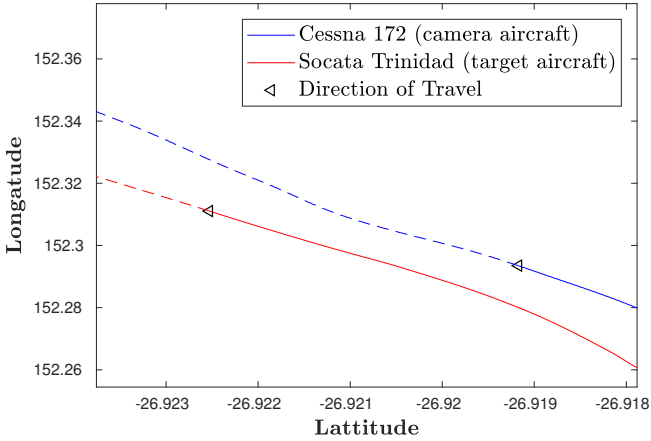}
  \caption{The flight path of the camera aircraft and target aircraft for Case S1. We note that there is an altitude separation between the aircraft of approximately $150$m.}
\label{fig:flightPath}
\end{center}
\end{figure}

We used $2$ tail-chase encounters designed to give the visual appearance of a true mid-air collision by minimising apparent motion with respect to the background. An example of the flight path for case S1 is presented in Figure \ref{fig:flightPath}.

\subsubsection{Moving Ground Vehicle Encounters (G1-G4)}
We used $4$ encounters containing moving cars on the road at the same time an aircraft was present in the field of view.

\subsubsection{Multiple Aircraft Encounter (M1)}
We used $1$  encounter in which a second (unknown) aircraft appeared in the field of view at the same time. We highlight that this second aircraft was not identified by the pilots at the time of the experiment and was only identified during post processing. We wanted to test the performance of our proposed system on this case to investigate whether it could effectively track multiple aircraft. 


\section{Experimental Results}\label{sec:results}
In this section we evaluate the performance of our proposed system on our head-on encounters,  stationary aircraft encounters,  moving ground vehicle encounters and our multiple aircraft encounter.

\subsection{Head-on Encounter Results}
Here we evaluate the performance of our proposed system on our head on encounters. We first characterise the performance of our system over a range of different window lengths $W$. We then examine the individual case performance. 

\subsubsection{System Operating Characteristic (SOC) Curve}
 \begin{figure}[]
\begin{center}
\includegraphics[scale=0.6]{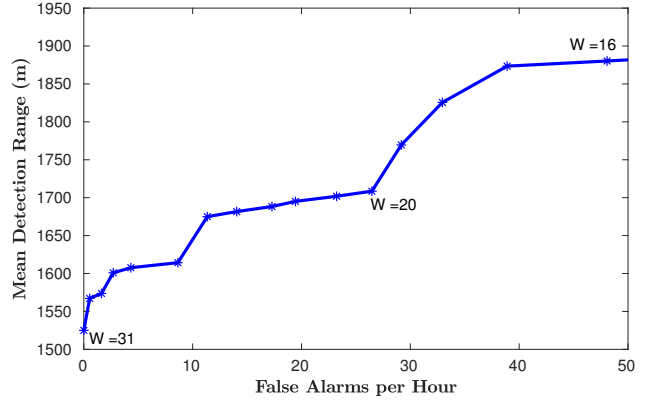}
  \caption{The mean detection ranges and false alarms per hour of our proposed  system for the head-on encounters overa range of different window lengths $W$. }
\label{fig:socCuve}
\end{center}
\end{figure}

The detection range and false alarm rate of our proposed system  varies with the choice of window length $W$  (sequential frames that we require the aircraft to be present for). 
To investigate the tradeoff between  detection range  and false alarm rate we composed system operating characteristic (SOC) curves which are commonly used  to evaluate the performance of a vision-based aircraft detection system \cite{Lai2013,molloy17,JamesRAL}.  
Figure \ref{fig:socCuve} presents the mean detection range versus the mean false alarms per hour for a range of different $W$ (shown by the $*$). For $0$ false alarms per hour the window length $W=31$ and for  $48$ false alarms per hour the window length $W=16$. The maximum standard error of mean (SEOM) of the mean detection ranges is $160$m false alarms per hour. 
Unsurprisingly an increase in detection ranges corresponds to an increase in the false alarm rate.

\subsubsection{Zero False Alarm (ZFA) Results}
\begin{figure}[]
\begin{center}
\includegraphics[scale=0.55]{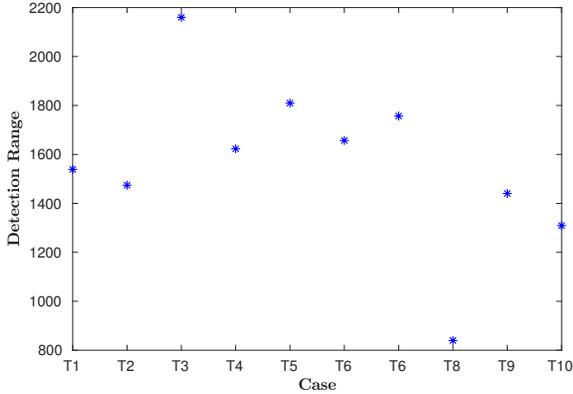}
  \caption{The individual ZFA detection ranges of our proposed  system for the head-on encnouters. The mean  detection range and SEOM are $1560$m and $109$m.}
\label{fig:detectionRanges}
\end{center}
\end{figure}

To examine the detection range of the individual cases we selected the minimum window length $W$ that achieves zero false alarms (ZFAs) ($W=31$) so we can examine the ZFA detection range as done in \cite{molloy17,Lai2013,JamesRAL}. 
Figure \ref{fig:detectionRanges} presents the individual detection ranges for T1-T10. The mean detection range and SEOM are $1560$m and $109$m respectively. 


\subsection{Stationary Aircraft Results}
\begin{figure}[]
\begin{center}
\includegraphics[scale=0.65]{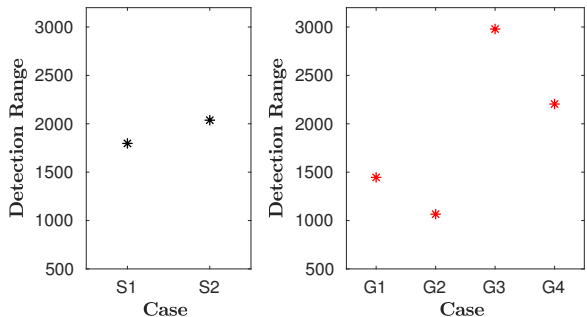}
  \caption{The individual ZFA detection ranges of our proposed  system for the stationary aircraft encounters (left) and the encounters with aircraft and rejection of moving ground vehicles (right). }
\label{fig:movingGround}
\end{center}
\end{figure}

We next evaluate the performance of our proposed system on our stationary aircraft cases.  As these cases were tail-chase encounters the target aircraft started closer to our camera aircraft and gradually got further away. We ran these cases through our proposed system in reverse to simulate the visual appearance of an aircraft approaching.
We were successfully able to detect the stationary aircraft in both encounters.  With window length $W=15$ the ZFA detection ranges for the $2$ cases are presented in Figure \ref{fig:movingGround} (left). 
The mean detection range and  SEOM are $1972$m and $120$m respectively. 
 
\subsection{Moving Ground Vehicle Results}
 We now  evaluated the performance of our system on our moving ground vehicle cases where a car and aircraft are both (simultaneously) in the field of view. We were successfully able to reject cars in all cases. With window length $W=19$ the ZFA detection ranges for the $4$ cases are presented in Figure \ref{fig:movingGround} (right). The mean detection range and SEOM are $1923$m and $423$m respectively.

\subsection{Multiple Aircraft Results}
\begin{figure}[]
\begin{center}
\includegraphics[scale=0.65]{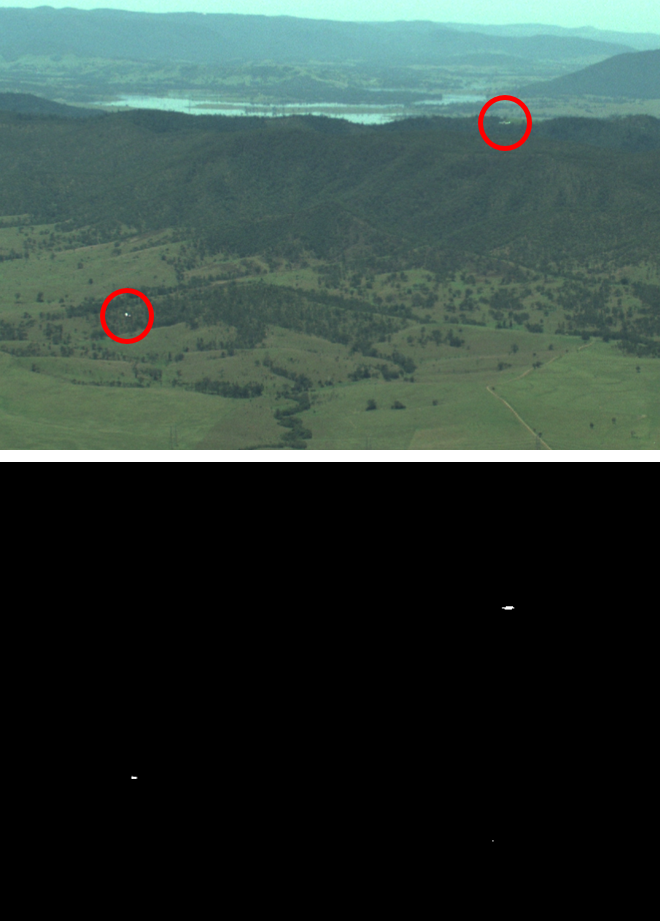}
  \caption{Our case with two aircraft present (top) the image pre-processing output where the white pixels represent potential aircraft (bottom). }
\label{fig:multi}
\end{center}
\end{figure}
Finally we evaluated the performance of our proposed system on our multiple aircraft case. We were able to detect the (known) aircraft at a ZFA range of $1720$m with window length $W = 22$.  Whilst our proposed network did  classify the third aircraft as seen in Figure \ref{fig:multi}, it would only classify the aircraft intermittently and hence was not detected by our proposed system (which requires an aircraft to be present in an image for  $W$ consecutive frames).

\section{Discussion}\label{sec:disc}
In this section we discuss our proposed system, its limitations and potential future work. For detection of aircraft below the horizon we proposed a multi-stage system which utilised deep learning as an image pre-processing stage combined with a simple temporal filtering stage which tracked potential aircraft over sequential frames.

We first evaluated the performance of our proposed system on a range of below horizon encounters. 
 We tested on 10 head-on encounters and were able to achieve a mean detection range and SEOM of $1560$m and $109$m respectively.
We then evaluated the performance of our system on 2 tail chase encounters where the aircraft exhibits small or no relative motion with respect to the background. We were able to successfully detect on both these cases with mean detection range and  SEOM of $1972$m and $120$m respectively. 
We next tested our proposed system on 4 cases with aircraft and cars both in the field of view (simultaneously). We were able to reject all  cars and detect the aircraft with a mean detection range and SEOM of $1923$m and $423$m respectively. 
Finally, we tested our system on a multiple aircraft case and whilst our network occasionally classified the aircraft correctly it was not detected by our our system (as it was not sequentially classified). This is potentially a results of bias in our training data as our data only consisted of head-on encounters and this was a crossing.

The detection ranges of our proposed system are comparable to the mean detection ranges reported in \cite{molloy17} ($1890$m with a SEOM of $43$) and in \cite{Nussberger2014} ($1714$ with an average of $4$ false alarms). Importantly our proposed system demonstrated new capabilities for below horizon aircraft detection including detection of stationary aircraft and rejection of moving ground vehicles. 

A key limitation of our proposed approach is the computational burden of the SegNet compared to existing frame differencing approaches. Prior SAA systems that have been implemented in Nvidia CUDA/C++ on GPUs performed at $15$ frames per second \cite{Lai2013}, $12$ frames per second \cite{molloy17} and $9$ frames per second \cite{Bratanov2017}.  Whilst our current  MATLAB implementation runs at $2$ frames per second, we expect the computational performance of our approach to improve with a specialised implementation.

A fundamental limitation in this application is the current lack of available aircraft data.  Image sequences which depict aircraft on true or near collision course are very limited and difficult to capture due to the risk of flying aircraft on converging paths \cite{lai2011}. Ideally we would have more training and testing data to encompass a range of aircraft and a range of scenarios. 

Future work could look at combining the system with a better temporal filtering stage (M/N frames rule, HMM filter or Kalman filter). Moreover our intuition is  that this system could be combined with \cite{JamesRAL} to design a unified system that worked for detection of aircraft above and below the horizon.

\section{Conclusion}\label{sec:conc}
In this paper we proposed a system for below horizon vision-based aircraft detection which exploited aircraft features rather than aircraft motion.  
Our proposed system was able to detect with comparable detection ranges and false alarm rates relative to the state of the art. Moreover our system addressed some of the key  challenges faced by the current state of the art including detection of aircraft with small or no relative motion to the background, rejection of moving ground vehicles and  multiple aircraft tracking.



\bibliographystyle{IEEEtran}
\bibliography{IEEEabrv,ref}

\begin{thebibliography}{10}
\providecommand{\url}[1]{#1}
\csname url@samestyle\endcsname
\providecommand{\newblock}{\relax}
\providecommand{\bibinfo}[2]{#2}
\providecommand{\BIBentrySTDinterwordspacing}{\spaceskip=0pt\relax}
\providecommand{\BIBentryALTinterwordstretchfactor}{4}
\providecommand{\BIBentryALTinterwordspacing}{\spaceskip=\fontdimen2\font plus
\BIBentryALTinterwordstretchfactor\fontdimen3\font minus
  \fontdimen4\font\relax}
\providecommand{\BIBforeignlanguage}[2]{{%
\expandafter\ifx\csname l@#1\endcsname\relax
\typeout{** WARNING: IEEEtran.bst: No hyphenation pattern has been}%
\typeout{** loaded for the language `#1'. Using the pattern for}%
\typeout{** the default language instead.}%
\else
\language=\csname l@#1\endcsname
\fi
#2}}
\providecommand{\BIBdecl}{\relax}
\BIBdecl

\bibitem{Researchandmarkets}
\BIBentryALTinterwordspacing
Researchandmarkets.com, ``{Unmanned Aerial Vehicle (UAV) Market to 2025 -
  Global Analysis and Forecasts by Component by Type and Application},'' Tech.
  Rep., 2018. [Online]. Available:
  \url{https://www.researchandmarkets.com/research/vx2jd5/}
\BIBentrySTDinterwordspacing

\bibitem{Clothier2015}
R.~A. Clothier, B.~P. Williams, and N.~L. Fulton, ``{Structuring the safety
  case for unmanned aircraft system operations in non-segregated airspace},''
  \emph{Safety Science}, vol.~79, pp. 213--228, Nov 2015.

\bibitem{Mcfadyen2016}
A.~Mcfadyen and L.~Mejias, ``{A survey of autonomous vision-based See and Avoid
  for Unmanned Aircraft Systems},'' \emph{Progress in Aerospace Sciences},
  vol.~80, pp. 1--17, 2016.

\bibitem{Nussberger2014}
A.~Nussberger, H.~Grabner, and L.~{Van Gool}, ``{Aerial object tracking from an
  airborne platform},'' in \emph{2014 International Conference on Unmanned
  Aircraft Systems (ICUAS)}.\hskip 1em plus 0.5em minus 0.4em\relax IEEE, May
  2014, pp. 1284--1293.

\bibitem{Lai2013}
J.~Lai, J.~J. Ford, L.~Mejias, and P.~O'Shea, ``{Characterization of sky-region
  morphological-temporal airborne collision detection},'' \emph{Journal of
  Field Robotics}, vol.~30, no.~2, pp. 171--193, Mar 2013.

\bibitem{zarandy2015novel}
A.~Zarandy, T.~Zsedrovits, B.~Pencz, M.~Nameth, and B.~Vanek, ``A novel
  algorithm for distant aircraft detection,'' in \emph{Unmanned Aircraft
  Systems (ICUAS), 2015 International Conference on}.\hskip 1em plus 0.5em
  minus 0.4em\relax IEEE, 2015, pp. 774--783.

\bibitem{JamesRAL}
J.~James, J.~J. Ford, and T.~L. Molloy, ``Learning to detect aircraft for long
  range, vision-based sense and avoid systems,'' \emph{IEEE Robotics and
  Automation Letters}, vol.~3, no.~4, pp. 4383 --4390, Aug 2018.

\bibitem{molloy17}
T.~L. Molloy, J.~J. Ford, and L.~Mejias, ``Detection of aircraft below the
  horizon for vision-based detect and avoid in unmanned aircraft systems,''
  \emph{Journal of Field Robotics}, vol.~34, no.~7, pp. 1378--1391.

\bibitem{Dey2011}
D.~Dey, C.~Geyer, S.~Singh, and M.~Digioia, ``{A cascaded method to detect
  aircraft in video imagery},'' \emph{The International Journal of Robotics
  Research}, vol.~30, no.~12, pp. 1527--1540, Oct 2011.

\bibitem{Rozantsev2014}
A.~Rozantsev, V.~Lepetit, and P.~Fua, ``Flying objects detection from a single
  moving camera,'' in \emph{2015 IEEE Conference on Computer Vision and Pattern
  Recognition (CVPR)}, Jun 2015, pp. 4128--4136.

\bibitem{Hwang2018}
S.~Hwang, J.~Lee, H.~Shin, S.~Cho, and D.~H. Shim, ``{Aircraft Detection using
  Deep Convolutional Neural Network in Small Unmanned Aircraft Systems},'' in
  \emph{2018 AIAA Information Systems-AIAA Infotech @ Aerospace}.\hskip 1em
  plus 0.5em minus 0.4em\relax Reston, Virginia: American Institute of
  Aeronautics and Astronautics, Jan 2018.

\bibitem{Badrinarayanan}
V.~Badrinarayanan, A.~Kendall, and R.~Cipolla, ``Segnet: A deep convolutional
  encoder-decoder architecture for image segmentation,'' \emph{IEEE
  Transactions on Pattern Analysis and Machine Intelligence}, vol.~39, no.~12,
  pp. 2481--2495, Dec 2017.

\bibitem{He2015}
K.~He, X.~Zhang, S.~Ren, and J.~Sun, ``Delving deep into rectifiers: Surpassing
  human-level performance on imagenet classification,'' in \emph{2015 IEEE
  International Conference on Computer Vision (ICCV)}, Dec 2015, pp.
  1026--1034.

\bibitem{Bratanov2017}
D.~Bratanov, L.~Mejias, and J.~J. Ford, ``{A vision-based sense-and-avoid
  system tested on a ScanEagle UAV},'' in \emph{2017 International Conference
  on Unmanned Aircraft Systems (ICUAS)}.\hskip 1em plus 0.5em minus 0.4em\relax
  IEEE, Jun 2017, pp. 1134--1142.

\bibitem{lai2011}
J.~Lai, L.~Mejias, and J.~J. Ford, ``{Airborne vision-based collision-detection
  system},'' \emph{Journal of Field Robotics}, vol.~28, no.~2, pp. 137--157,
  Mar 2011.

\end{thebibliography}


\end{document}